\documentclass[11pt,twocolumn,twoside]{article}

\usepackage[T1]{fontenc}
\usepackage[utf8]{inputenc}
\usepackage{lmodern}
\usepackage{microtype}
\usepackage{abstract} % in preamble

\usepackage{amsmath,amssymb}
\usepackage{enumitem}   % for [leftmargin=*] in itemize

\usepackage[margin=1in]{geometry}
\usepackage{graphicx}
\usepackage{subcaption} % for arranging subfigures
\usepackage{booktabs}
\usepackage{siunitx}
\usepackage{dblfloatfix} % robust two-column float placement (top & bottom)
\usepackage{placeins}    % \FloatBarrier to stop floats escaping sections
\graphicspath{{figures/}}

\usepackage{hyperref}
\usepackage[nameinlink,noabbrev]{cleveref} % keep after hyperref
\hypersetup{colorlinks=true, linkcolor=black, citecolor=black, urlcolor=black}
\title{Gait Recognition with Temporal Kolmogorov--Arnold Networks}
\date{}

\author{
Mohammed Asad\textsuperscript{1} \;{\large\textbullet}\; Dinesh Kumar Vishwakarma\textsuperscript{1}\\[2pt]
\textsuperscript{1}Multi-Modal Data Analytics Lab, Department of Information Technology\\
Delhi Technological University, Delhi, 110042, India\\
\href{mailto:asadmohd9411@gmail.com}{asadmohd9411@gmail.com} \;{\small\textbullet}\;
\href{mailto:dinesh@dtu.ac.in}{dinesh@dtu.ac.in}
}

\begin{document}

% =========================================================
% Abstract
% =========================================================
\twocolumn[
\maketitle
\vspace{-0.75em}
\begin{onecolabstract}
Gait recognition is a biometric modality that identifies individuals from their characteristic walking patterns. Unlike conventional biometric traits, gait can be acquired at a distance and without active subject cooperation, making it suitable for surveillance and public safety applications. Nevertheless, silhouette-based temporal models remain sensitive to long sequences, observation noise, and appearance-related covariates. Recurrent architectures often struggle to preserve information from earlier frames and are inherently sequential to optimize, whereas transformer-based models typically require greater computational resources and larger training sets and may be sensitive to irregular sequence lengths and noisy inputs. These limitations reduce robustness under clothing variation, carrying conditions, and view changes, while also hindering the joint modeling of local gait cycles and longer-term motion trends. To address these challenges, we introduce a Temporal Kolmogorov--Arnold Network (TKAN) for gait recognition. The proposed model replaces fixed edge weights with learnable one-dimensional functions and incorporates a two-level memory mechanism consisting of short-term RKAN sublayers and a gated long-term pathway. This design enables efficient modeling of both cycle-level dynamics and broader temporal context while maintaining a compact backbone. Experiments on the CASIA-B dataset indicate that the proposed CNN+TKAN framework achieves strong recognition performance under the reported evaluation setting.
\end{onecolabstract}
\vspace{0.5em}
]

% =========================================================
% 1. Introduction
% =========================================================
\section{Introduction}
Reliable identification from CCTV video is difficult because faces are often low resolution, off-angle, or occluded. Gait recognition offers a practical alternative: it leverages how a person walks and works at a distance without cooperation. However, real scenes introduce significant variability: cross-view changes warp silhouettes; shadows, motion blur, and background leakage from imperfect segmentation corrupt contours; and covariates such as bags or clothing changes alter both appearance and motion. Irregular frame rates, short or missing sequences, and view gaps also disrupt the periodic yet deformable walking pattern. Surveys document these as dominant failure modes for silhouette-based systems especially on CASIA–B under BG and CL where view, clothing, and carrying conditions consistently reduce accuracy \cite{Shen2022Survey,Munusamy2024Survey}. The core challenge is therefore not collecting more video, but designing temporal models that remain reliable under noisy silhouettes, changing viewpoints, and appearance shifts.

Recent progress has targeted stronger spatial and temporal representations. Silhouette pipelines introduced part-aware micro-motion (GaitPart)~\cite{Fan2020GaitPart}, which splits the silhouette into body regions (e.g., legs, torso, arms) and tracks short-range motions per part to capture fine swings and shifts; global–local coupling (GaitGL)~\cite{Lin2021GaitGL}, which learns whole-body shape and local details together and lets them inform each other; and context-sensitive multi-scale temporal fusion (CSTL)~\cite{Huang2021CSTL}, which fuses short/medium/long time windows with clip-dependent weights. 3D local operations~\cite{Huang2021Local3D} apply local 3D convolutions on adaptive space–time neighborhoods matched to part size and motion rhythm. Beyond silhouettes, skeleton graphs (GaitGraph)~\cite{Teepe2021GaitGraph} represent joints and bones as a graph to model pose dynamics with less sensitivity to clothing/background, while optical-flow pyramids (GOFN)~\cite{Ye2023GOFN} capture multi-scale pixel motion to complement appearance cues. Despite this progress, temporal heads still face trade-offs: recurrent models can suffer vanishing gradients and learning from events that happened many steps earlier, Transformers often need larger data and compute and can be brittle to sequence noise and length; and simple temporal pooling underfits the variability induced by view and clothing. What is missing is a temporal module that is expressive across gait cycles, stable and data efficient on standard benchmarks, and lightweight enough that gains do not come from a larger backbone.

We propose a compact spatiotemporal framework that combines a lightweight frame-wise convolutional encoder with a Temporal Kolmogorov--Arnold Network (TKAN). Kolmogorov--Arnold Networks (KANs) replace fixed edge weights with learnable univariate functions, providing a smooth functional parameterization with an inductive bias distinct from that of conventional multilayer perceptrons \cite{Liu2024KAN}. TKAN extends this formulation to sequential data through a two-level memory design in which short-term Recurrent KAN (RKAN) sublayers capture cycle-scale variations and a gated long-term pathway preserves broader temporal context \cite{Genet2024TKAN}. In the proposed pipeline, a CNN first extracts a compact representation from each frame, after which the TKAN head models both local gait dynamics and longer-range temporal structure. The resulting sequence representation is then temporally aggregated and passed to a classifier. All comparative models employ the same CNN backbone and optimization setting so that performance differences can be attributed to the temporal head. Evaluation follows the standard CASIA-B subject protocol (IDs 1--74 for training and 75--124 for testing, with identical-view matches excluded) \cite{CASIAB}.

The principal contributions of this work are as follows:\\
(1) We show that a TKAN-based temporal head models periodic yet deformable gait more effectively by combining edge-wise learnable functions with a two-scale memory mechanism, while remaining lightweight. \\
(2) We provide a controlled comparison with LSTM- and Transformer-based temporal heads under identical encoder capacity and optimization settings, allowing the effect of the temporal module to be isolated.\\ 
(3) On CASIA-B, the proposed CNN+TKAN model achieves Rank-1 accuracies of 99.52\% under NM, 99.56\% under BG, and 98.82\% under CL. These gains, especially under CL, support the value of multi-timescale temporal modeling beyond a single operating point.

% =========================================================
% 2. Related Work
% =========================================================
\section{Related Work}
\label{sec:related}
Silhouette-based deep models are widely used for gait recognition. \href{https://openaccess.thecvf.com/content_CVPR_2020/html/Fan_GaitPart_Temporal_Part-Based_Model_for_Gait_Recognition_CVPR_2020_paper.html}{GaitPart}~\cite{Fan2020GaitPart} splits a silhouette into body regions (e.g., legs, torso, arms) and applies local temporal operations per region to capture micro-motions like leg swings and arm sway. \href{https://openaccess.thecvf.com/content/ICCV2021/html/Lin_Gait_Recognition_via_Effective_Global-Local_Feature_Representation_and_Local_Temporal_ICCV_2021_paper.html}{GaitGL}~\cite{Lin2021GaitGL} couples a global branch for overall shape with a local branch for fine details via GLConv, so that both scales reinforce each other. \href{https://openaccess.thecvf.com/content/ICCV2021/papers/Huang_Context-Sensitive_Temporal_Feature_Learning_for_Gait_Recognition_ICCV_2021_paper.pdf}{CSTL}~\cite{Huang2021CSTL} fuses short-, medium-, and long-range temporal windows using clip-dependent weights that adapt to context. \href{https://openaccess.thecvf.com/content/ICCV2021/html/Huang_3D_Local_Convolutional_Neural_Networks_for_Gait_Recognition_ICCV_2021_paper.html}{3D Local CNNs}~\cite{Huang2021Local3D} introduce 3D local convolution blocks that operate on adaptive space–time neighborhoods matched to part size and motion rhythm. GLN~\cite{Hou2020GLN} shows that lateral aggregation with hierarchical pooling can produce compact, highly discriminative silhouette descriptors.

A second line explicitly models temporal dependencies with recurrence or attention on silhouettes. ConvLSTM over frame-by-frame gait images learns spatio-temporal patterns directly from silhouettes without explicit gait-cycle segmentation~\cite{ConvLSTMffGEI2020}. CNN+BiLSTM pairs a per-frame encoder with a bidirectional LSTM so past and future frames jointly refine the sequence representation~\cite{ConvBiLSTM2021}. Long–short-term attention (LSTA) replaces recurrence with attention windows that emphasize both local (few frames) and global (whole clip) relations~\cite{LSTA2022}. Vision Transformer for gait (Gait-ViT) treats a gait image/sequence as patches and uses self-attention to capture long-range structure~\cite{GaitViT2022}. Multi-Scale Context-Aware Transformers (MCAT) extract short/medium/long temporal cues and fuse them with local/global self-attention for robust aggregation~\cite{MCAT2023}. Taken together, these studies improve either local motion encoding or long-range aggregation, but they still leave open how to capture both cycle-level detail and longer-range context under covariate shift.

Robustness to confounders is addressed from complementary angles. GaitGCI~\cite{Dou2023GaitGCI} uses generative counterfactual intervention to suppress nuisance regions and highlight identity-relevant evidence. CLTD~\cite{Xiong2024CLTD} applies causality-inspired discriminative learning across spatial, temporal, and spectral domains to disentangle identity from view, clothing, and carrying effects. VPNet~\cite{Ma2024VPNet} learns visual prompts and a dynamic transformer to inject task priors into the temporal encoder. Beyond silhouettes, skeleton graphs~\cite{Teepe2021GaitGraph,Fan2024SkeletonGait} model joints and bones to reduce sensitivity to clothing/background; SMPLGait~\cite{Zheng2022SMPLGait} fuses silhouettes with SMPL-based 3D shape and viewpoint cues; and optical-flow pyramids like GOFN~\cite{Ye2023GOFN} capture multi-scale pixel motion to complement appearance cues.

Our work stays with silhouettes but replaces conventional RNN/Transformer heads with a Temporal Kolmogorov–Arnold Network (TKAN), which uses edge-wise learnable univariate functions and a two-tier memory to model periodic yet deformable sequence dynamics while remaining lightweight and data-efficient.

% =========================================================
% 3. Proposed Methodology
% =========================================================
\section{Proposed Methodology}\label{sec:method}

\subsection{Overview and Workflow}
We consider a compact silhouette-based gait recognition pipeline composed of a frame-wise convolutional encoder, a temporal modeling module, and an identity classifier. Let $\{I_t\}_{t=1}^{T}$ denote grayscale silhouette frames resized to $64\times64$. A CNN maps each frame to a feature vector $x_t \in \mathbb{R}^{256}$, yielding the sequence $X=[x_1,\dots,x_T]\in\mathbb{R}^{T\times256}$. The temporal module then aggregates $X$ into an identity posterior over the $C$ training classes. To ensure a controlled comparison, we evaluate three temporal heads, namely LSTM, Transformer, and the proposed Temporal Kolmogorov--Arnold Network (TKAN), while keeping both the CNN backbone and the optimization setting unchanged. Under this design, performance differences can be attributed directly to the temporal modeling component.

% ---- Figure 1: two-column (starred) ----
\begin{figure*}[!tbp]
  \centering
  \includegraphics[width=\textwidth]{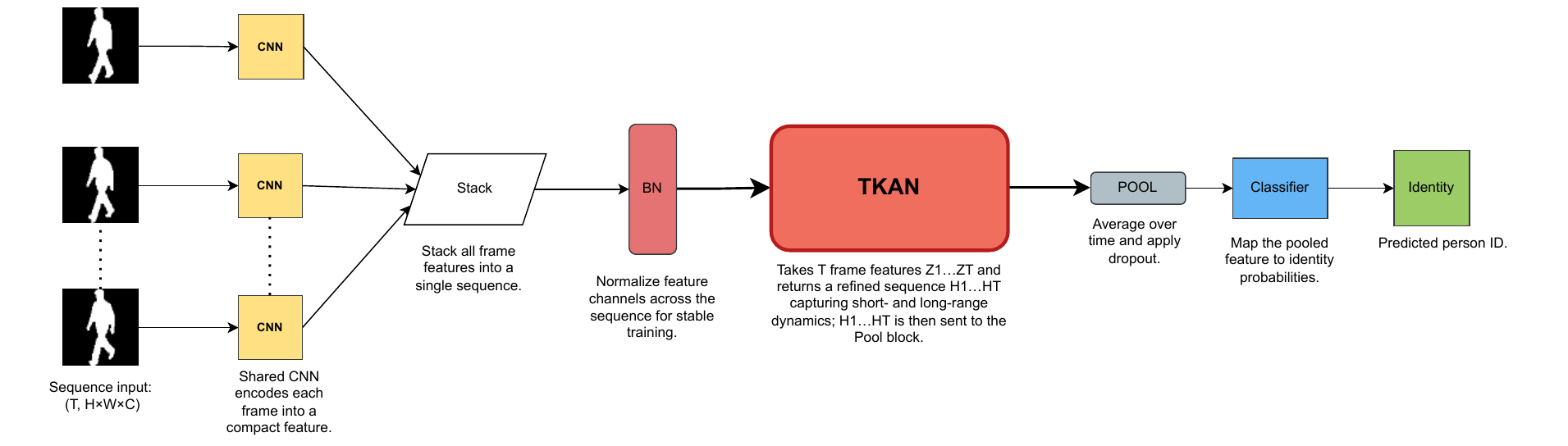}
  \caption{Proposed pipeline. A shared CNN encodes each silhouette frame into a compact feature; the frame features are stacked and batch-normalized to form a sequence. The TKAN head models short- and long-range gait dynamics over the sequence and outputs refined time-step features, which are temporally averaged (with dropout) to obtain a clip-level descriptor. A linear–softmax classifier maps this descriptor to the identity posterior and the top-1 label is reported. (In our setup: $T{=}50$, feature width $d{=}256$, classes $C{=}74$.)}
  \label{fig:workflow}
\end{figure*}

\subsection{Frame-wise CNN Encoder}\label{sec:cnn}
\begin{figure}[t]
  \centering
  \includegraphics[width=\linewidth]{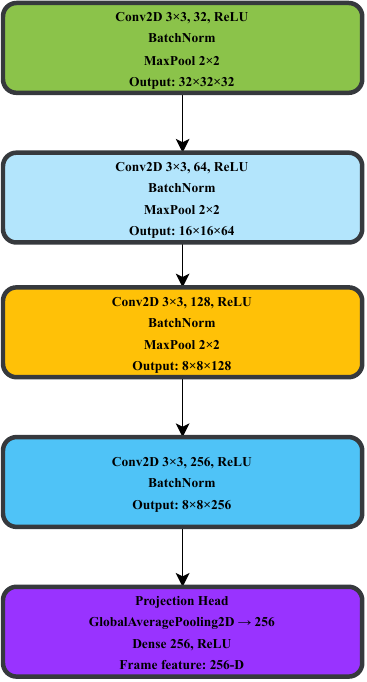}
  \caption{Architecture of CNN model}
  \label{fig:cnn}
\end{figure}

We use the same compact CNN for all models (LSTM, Transformer, TKAN) so performance differences come only from the temporal module. The layout is shown in Figure~\ref{fig:cnn}.

\begin{itemize}
  \setlength\itemsep{0.3em}
  \item \textbf{Input:} single grayscale silhouette frame (64$\times$64$\times$1).
  \item \textbf{Block 1:} 3$\times$3 conv (32) $\rightarrow$ BatchNorm $\rightarrow$ 2$\times$2 max pool. Learns low-level edges and reduces spatial size.
  \item \textbf{Block 2:} 3$\times$3 conv (64) $\rightarrow$ BatchNorm $\rightarrow$ 2$\times$2 max pool. Builds mid-level shape cues with stable training.
  \item \textbf{Block 3:} 3$\times$3 conv (128) $\rightarrow$ BatchNorm $\rightarrow$ 2$\times$2 max pool. Encodes part-level patterns of legs, arms, and torso.
  \item \textbf{Block 4:} 3$\times$3 conv (256) $\rightarrow$ BatchNorm $\rightarrow$ Global Average Pool. Summarizes each channel into a shift-tolerant descriptor.
  \item \textbf{Projection:} fully connected layer (256, ReLU). Produces a compact 256-D feature per frame.
  \item \textbf{Sequence use:} the CNN runs on every frame (time-distributed), yielding a $T \times 256$ sequence that the temporal module aggregates.
\end{itemize}

% ---- Figure 2: single-column ----

\subsection{Temporal Kolmogorov--Arnold Networks (TKAN)}
\label{sec:tkan}

\textbf{Kolmogorov--Arnold Networks (KAN).}
KAN are motivated by the Kolmogorov--Arnold representation theorem, which states that any continuous multivariate function $f:\,[0,1]^n\!\to\!\mathbb{R}$ can be expressed as a superposition of continuous univariate functions and addition:
\begin{equation}
f(x_1,\ldots,x_n)
\;=\;
\sum_{q=1}^{2n+1}
\Phi_q\!\left(
\sum_{p=1}^{n}\phi_{q,p}(x_p)
\right),
\label{eq:KAT}
\end{equation}
where $\{\phi_{q,p}\}$ and $\{\Phi_q\}$ are continuous scalar functions.
KAN operationalize \eqref{eq:KAT} by replacing fixed weights in a layer with learnable univariate maps placed on edges. 
Let $x^{(\ell)}\!\in\!\mathbb{R}^{n_\ell}$ denote the activations at layer $\ell$ and $x^{(\ell+1)}\!\in\!\mathbb{R}^{n_{\ell+1}}$ at layer $\ell{+}1$.
Each edge $(i\!\to\!j)$ carries a function $\varphi^{(\ell)}_{j,i}:\mathbb{R}\!\to\!\mathbb{R}$, and the layer computes
\begin{align}
z^{(\ell+1)}_j &= \sum_{i=1}^{n_\ell} \varphi^{(\ell)}_{j,i}\!\big(x^{(\ell)}_i\big),
\qquad j=1,\ldots,n_{\ell+1}, \label{eq:kan-sum} \\
x^{(\ell+1)}_j &= \rho\!\big(z^{(\ell+1)}_j\big), \label{eq:kan-act}
\end{align}
with a pointwise nonlinearity $\rho(\cdot)$.
Each $\varphi^{(\ell)}_{j,i}$ is parameterized as a smooth 1D basis expansion with a linear skip,
\begin{equation}
\varphi^{(\ell)}_{j,i}(u) \;=\;
\alpha^{(\ell)}_{j,i}\,u \;+\; \sum_{k=1}^{K} a^{(\ell)}_{j,i,k}\, b_k(u),
\label{eq:kan-edge}
\end{equation}
where $\{b_k\}$ are fixed bases (e.g., B-splines/RBFs) and $\{\alpha^{(\ell)}_{j,i},a^{(\ell)}_{j,i,k}\}$ are learned coefficients.
Stacking layers gives a deep KAN:
\begin{equation}
\mathrm{KAN}(x)=\big(\Phi^{(L-1)}\circ\cdots\circ\Phi^{(0)}\big)(x),
\label{eq:kan-deep}
\end{equation}
where each $\Phi^{(\ell)}$ is the function matrix induced by \eqref{eq:kan-sum}–\eqref{eq:kan-edge}.
By learning edge-wise univariate functions, KANs provide compact and smooth approximations while preserving the usual tensor I/O.

\medskip
\textbf{Temporal KAN (TKAN).}
TKAN models gait with two complementary pathways: a short-term RKAN branch that captures cycle-level motion and a long-term gated branch that preserves information across many frames, with local RKAN responses modulating the final output.

Let a feature sequence be $X=\{x_t\}_{t=1}^{T}$, $x_t\in\mathbb{R}^{d}$. 
For $m=1,\ldots,M$ parallel RKAN sub-layers, define
\begin{align}
s_{m,t} &= W_{m,\tilde{x}}\,x_t \;+\; W_{m,\tilde{h}}\,\tilde{h}_{m,t-1} \;+\; b_m, \label{eq:rkan-pre} \\
\tilde{o}_{m,t} &= \phi_m\!\big(s_{m,t}\big), \qquad \text{(KAN block as in \eqref{eq:kan-edge})} \label{eq:rkan-phi} \\
\tilde{h}_{m,t} &= W_{m,hh}\,\tilde{h}_{m,t-1} \;+\; W_{m,hz}\,\tilde{o}_{m,t}, \label{eq:rkan-state}
\end{align}
where $\tilde{h}_{m,t}$ is the short-term state of sub-layer $m$ and $\phi_m$ is a learnable edge-function module.
The RKAN responses are concatenated as $r_t=\mathrm{Concat}[\tilde{o}_{1,t},\ldots,\tilde{o}_{M,t}]$.
The long-term gated pathway then uses $r_t$ to control the output gate:

% --- SPLIT into compact single-column equation blocks (labels preserved) ---
\begin{equation}
\begin{aligned}
f_t &= \sigma(W_f x_t + U_f h_{t-1} + b_f),\\
i_t &= \sigma(W_i x_t + U_i h_{t-1} + b_i),
\end{aligned}
\label{eq:tkan-fi}
\end{equation}

\begin{equation}
\begin{aligned}
\tilde{c}_t &= \tanh(W_c x_t + U_c h_{t-1} + b_c),\\
o_t &= \sigma(W_o r_t + b_o),
\end{aligned}
\label{eq:tkan-oc}
\end{equation}

\begin{equation}
\begin{aligned}
c_t &= f_t \odot c_{t-1} + i_t \odot \tilde{c}_t,\\
h_t &= o_t \odot \tanh(c_t).
\end{aligned}
\label{eq:tkan-core}
\end{equation}

Thus TKAN fuses fast, function-parameterized RKAN responses with a slow, gated memory that preserves longer context.

\medskip
\textbf{TKAN in this work (instantiation for gait).}
Each clip is processed into $T{=}50$ frame features with dimensionality $d{=}256$ from the shared CNN encoder, yielding $X\in\mathbb{R}^{T\times d}$. 
TKAN uses $M{=}2$ RKAN sub-layers \eqref{eq:rkan-pre}–\eqref{eq:rkan-state} in parallel and a $d$-dimensional gated path \eqref{eq:tkan-fi}–\eqref{eq:tkan-core}. 
Per-step states $\{h_t\}$ are aggregated by temporal average pooling to obtain a clip descriptor and, during training, are passed to a classifier:
\begin{equation}
\label{eq:tkan-cls}
\begin{aligned}
\bar{h} &= \frac{1}{T}\sum_{t=1}^{T} h_t \in \mathbb{R}^{d},\\
p(y\mid X) &= \mathrm{softmax}(W_y\,\bar{h}+b_y).
\end{aligned}
\end{equation}

Dropout (rate $0.3$) is applied before the final layer during training. The network is optimized with cross-entropy over the $C{=}74$ training identities:
\begin{equation}
\mathcal{L} \;=\; -\frac{1}{N}\sum_{n=1}^{N}\log p\!\big(y_n\mid X_n\big),
\label{eq:ce-loss}
\end{equation}
At inference time, the classifier layer is not used; instead, the pooled descriptor $\bar{h}$ serves as the gait embedding for gallery-probe matching on the test subjects. All compared temporal heads (LSTM, Transformer, and TKAN) use the same encoder, feature width $d$, and optimization setting so that the comparison isolates the effect of the temporal module.

\subsection{Temporal Baselines}
\textbf{LSTM:} Two LSTM layers are stacked in series. The first layer has 256 hidden units and returns a feature sequence at all time steps, while the second layer has 128 hidden units and also returns the full output sequence. The resulting sequence is temporally averaged, followed by dropout ($p{=}0.3$) and a softmax classifier.

\textbf{Transformer:} Two transformer encoder layers (4 attention heads, model width 256, feed-forward width 1024) with learned positional embeddings to encode order. After the encoders, we average across time (temporal pooling), apply dropout ($p{=}0.3$), and use a softmax classifier.

\subsection{Optimization Summary}
All models use the same training setup for a fair comparison: class-balanced cross-entropy on 74 training identities, Adam optimizer (learning rate $1\times10^{-4}$), and batch size $8$. We apply early stopping (patience $10$) and ReduceLROnPlateau (factor $0.5$, patience $5$). Each clip has $T{=}50$ frames, and frames are normalized to $[0,1]$.

% =========================================================
% 4. Experiments and Results
% =========================================================
\section{Experiments and Results}
\label{sec:experiments}

This section tests whether TKAN improves temporal modeling for silhouettes without changing the CNN backbone. For a fair comparison, all methods use the same frame-wise CNN and training setup; only the temporal module (LSTM, Transformer, or TKAN) differs. We follow the standard CASIA–B subject split (IDs 1–74 train, 75–124 test; identical-view excluded) and report Rank-1/Rank-5 identification and AUC trends, along with training curves to assess stability. We first describe the dataset and preprocessing, then summarize the shared optimization setting, and finally present the comparative results and training dynamics.

\subsection{Dataset}
\label{sec:dataset}
The proposed method is evaluated on the CASIA--B dataset, one of the most widely used benchmarks for gait recognition. The dataset contains 124 subjects, each with 10 gait sequences acquired under three walking conditions: 6 normal walking sequences (NM), 2 sequences recorded while carrying a bag (BG), and 2 sequences recorded while wearing a coat or bulky clothing (CL).

\subsubsection{Training--testing division}
\label{sec:split}
Following the standard CASIA--B evaluation protocol, subjects 1--74 are used for training and subjects 75--124 are reserved for testing.

During training, all sequences from the 74 training subjects are used to optimize the network as a closed-set classifier over the training identities. During testing, the final classification layer is discarded and the pooled descriptor $\bar{h}$ is used as the gait embedding. Following the standard CASIA-B LT protocol, for each test subject the NM\#01--04 sequences form the gallery set, whereas NM\#05--06, BG\#01--02, and CL\#01--02 form three probe sets. Similarity between probe and gallery embeddings is computed using cosine similarity, identical-view matches are excluded, and Rank-1 and Rank-5 accuracies are obtained by sorting gallery candidates according to similarity.

\subsection{Feature dimensions and data preparation}
\label{sec:features}
Binary silhouette images are converted to grayscale, resized to $64\times64$, and normalized to the range $[0,1]$. Each sequence is segmented or padded to a fixed length of $T{=}50$ frames. The shared frame-wise CNN encoder produces a 256-dimensional feature vector for each frame, resulting in a $T\times256$ feature sequence. The temporal module then aggregates this sequence into a clip-level descriptor. During training, dropout ($p{=}0.3$) and a softmax classifier are applied; during testing, the descriptor is used for similarity-based gallery-probe matching.

\subsection{Parameter Settings and Observations}
\label{sec:params}

All models are trained under an identical experimental configuration so that observed differences arise only from the temporal component. Optimization is performed using class-balanced cross-entropy, the Adam optimizer with a learning rate of $1\times10^{-4}$, and a batch size of 8. Early stopping is applied with a patience of 10 epochs, and the learning rate is adjusted using ReduceLROnPlateau with a factor of $0.5$ and a patience of 5. Each input clip contains $T{=}50$ frames. Silhouettes are converted to grayscale, resized to $64\times64$, normalized to $[0,1]$, and encoded frame-wise by the shared CNN to produce a $T\times256$ feature sequence for temporal modeling.

% ---- Training curves: wide two-column figure with controlled height
\begin{figure}[!t]
  \centering
  \includegraphics[width=\linewidth]{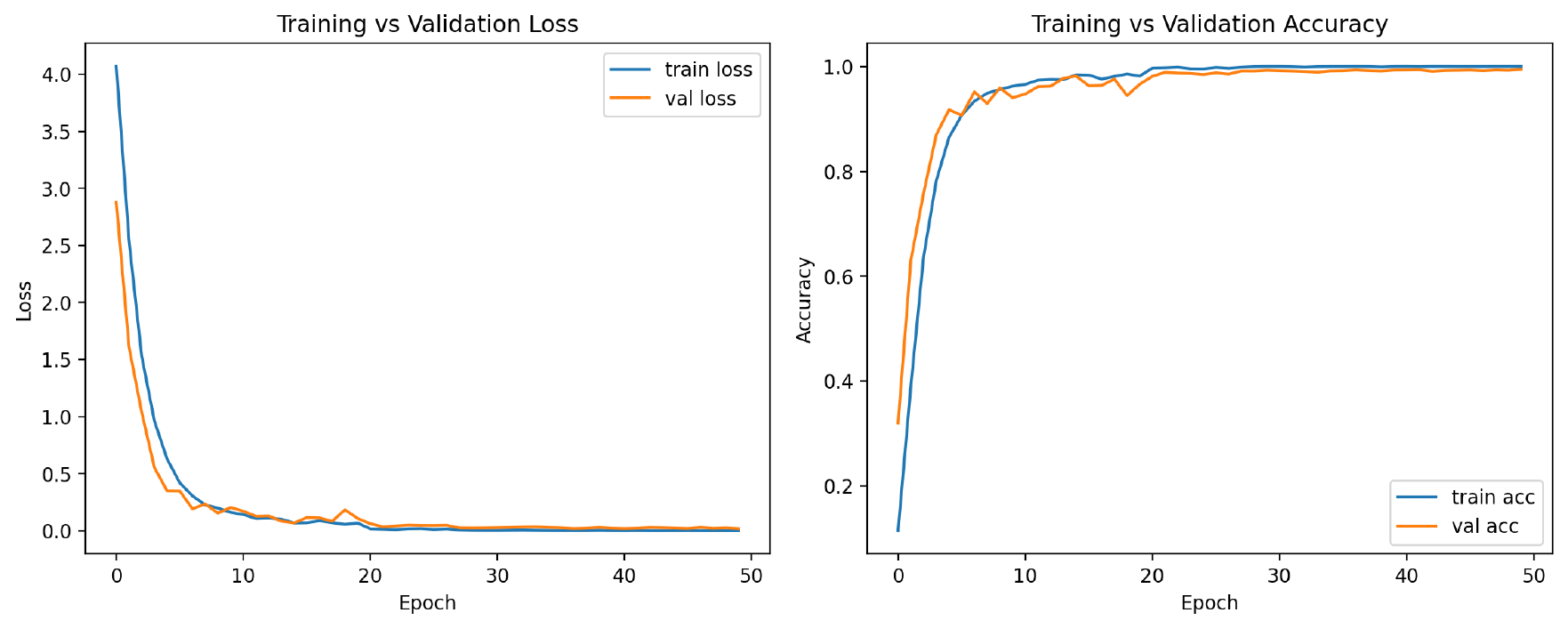}
  \caption{Training/validation loss and accuracy for CNN+TKAN on CASIA--B (subject split 1--74 train, 75--124 test).}
  \label{fig:curves}
\end{figure}

Under this common setting, CNN+TKAN exhibits stable optimization behaviour and no evidence of pronounced late-epoch overfitting, as shown in \cref{fig:curves}. Additional sweeps around the default configuration, with learning rates in the range $[5\times10^{-5},\,2\times10^{-4}]$ and dropout values in $[0.2,\,0.4]$, produce similar performance trends; the selected setting of learning rate $10^{-4}$ and dropout $0.3$ consistently yields the best validation and test results. To assess ranking quality beyond a single operating point, we report AUC-versus-epoch curves for NM, BG, and CL using both micro and macro averaging in \cref{fig:auc_curves}. The resulting curves are close to unity for NM and BG and show a consistent margin on CL, indicating strong threshold-independent separability.

\subsection{Results}
\label{sec:results}

\Cref{tab:casia-sota} summarizes Rank-1 accuracy on the CASIA--B dataset. Under the reported setting, CNN+TKAN attains the highest accuracy in all three evaluation conditions while retaining the same lightweight CNN backbone. Specifically, the model achieves 99.52\% on NM, 99.56\% on BG, and 98.82\% on CL. The largest gains appear under clothing variation, the most challenging condition, suggesting that TKAN is most useful when appearance cues are distorted and temporal continuity must carry more of the identity signal. Averaged across NM, BG, and CL, CNN+TKAN obtains 99.30\%, indicating strong robustness without increasing backbone complexity.

\begin{table}[!b]
\centering
\caption{Comparison on CASIA-B (Rank-1 \%) under the standard protocol (subjects 1--74 train, 75--124 test; identical-view excluded). \textsuperscript{\dag} indicates enlarged resolution (e.g., 128$\times$88).}
\label{tab:casia-sota}
\scriptsize
\setlength{\tabcolsep}{3pt}
\resizebox{\columnwidth}{!}{%
\begin{tabular}{lcccc}
\toprule
Method & NM & BG & CL & Avg \\
\midrule
\textbf{CNN+TKAN (Ours)} & \textbf{99.52} & \textbf{99.56} & \textbf{98.82} & \textbf{99.30} \\
\midrule
VPNet (CVPR'24)~\cite{Ma2024VPNet} & 98.3 & 96.3 & \textbf{90.0} & 94.9 \\
CLTD (ECCV'24)~\cite{Xiong2024CLTD} & \textbf{98.6} & \textbf{96.4} & 89.3 & 94.8 \\
HSTL (ICCV'23)~\cite{Wang2023HSTL} & 98.1 & 95.9 & 88.9 & 94.3 \\
GaitGCI (CVPR'23)~\cite{Dou2023GaitGCI} & 98.4 & 96.6 & 88.5 & 94.5 \\
DyGait (ICCV'23)~\cite{Wang2023DyGait} & 98.4 & 96.2 & 87.8 & 94.1 \\
GaitGS (ICIP'24)~\cite{Xiong2024GaitGS} & 97.9 & 94.6 & 86.9 & 93.1 \\
GaitSnippet (arXiv'25)~\cite{Hou2025GaitSnippet} & 97.8 & 95.7 & 86.7 & 93.4 \\
GaitSnippet\textsuperscript{\dag}~(arXiv'25)~\cite{Hou2025GaitSnippet} & \textbf{99.0} & \textbf{97.3} & 88.7 & \textbf{95.0} \\
GaitGS\textsuperscript{\dag}~(ICIP'24)~\cite{Xiong2024GaitGS} & 98.2 & 96.5 & \textbf{89.7} & 94.8 \\
\bottomrule
\end{tabular}%
}
\end{table}

With the encoder and optimization setting held constant, \Cref{tab:internals} compares LSTM, Transformer, and TKAN as alternative temporal heads. TKAN yields the best performance in all three conditions, improving over LSTM by 1.46 points on NM, 2.80 points on BG, and 6.26 points on CL, and improving over the Transformer baseline by 0.66, 2.06, and 3.61 points, respectively. The largest gains occur under clothing variation, suggesting that the proposed temporal formulation is especially effective in more challenging scenarios. Rank-5 accuracy remains close to 100\% for all methods, whereas the main benefit of TKAN is reflected in the improvement of Rank-1 accuracy.

\begin{table}[!b]
\centering
\caption{Internal comparison (same CNN, training, and protocol). Rank-1 / Rank-5 (\%) on CASIA-B by condition.}
\label{tab:internals}
\scriptsize
\setlength{\tabcolsep}{2pt}
\resizebox{\columnwidth}{!}{%
\begin{tabular}{lccc}
  \toprule
  \textbf{Method} & \textbf{NM} & \textbf{BG} & \textbf{CL} \\
  \midrule
  CNN + LSTM        & 98.06 / 99.63 & 96.76 / 99.56 & 92.56 / 98.90 \\
  CNN + Transformer & 98.86 / 99.82 & 97.50 / 99.85 & 95.21 / 98.45 \\
  \textbf{CNN + TKAN (ours)}
                     & \textbf{99.52} / \textbf{100.00}
                     & \textbf{99.56} / \textbf{100.00}
                     & \textbf{98.82} / \textbf{100.00} \\
  \bottomrule
\end{tabular}%
}
\end{table}

% ---- Figure 4 follows the table discussion and may float to the next page top

\begin{figure*}[!t]
  \centering
  \begin{subfigure}[t]{0.32\textwidth}
    \centering
    \includegraphics[width=\linewidth]{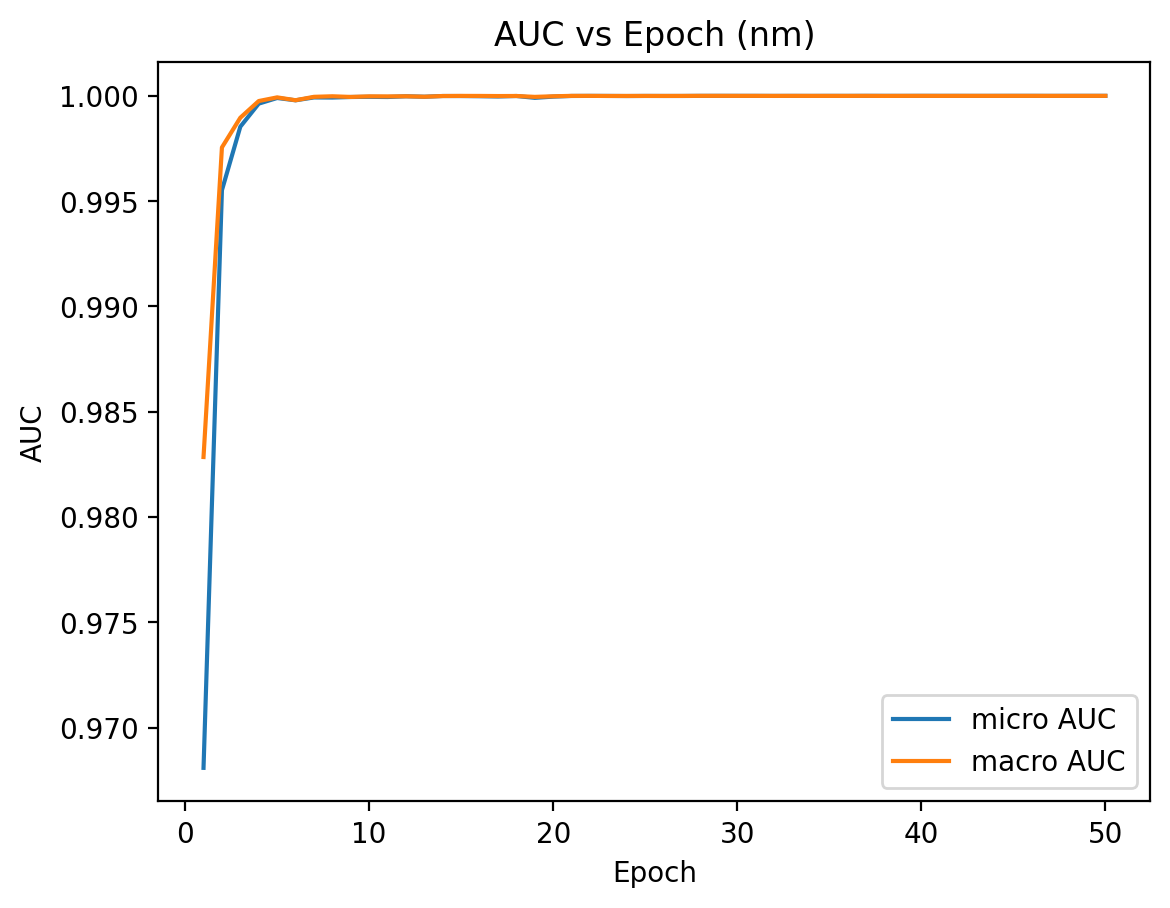}
    \caption*{\small NM}
  \end{subfigure}\hfill
  \begin{subfigure}[t]{0.32\textwidth}
    \centering
    \includegraphics[width=\linewidth]{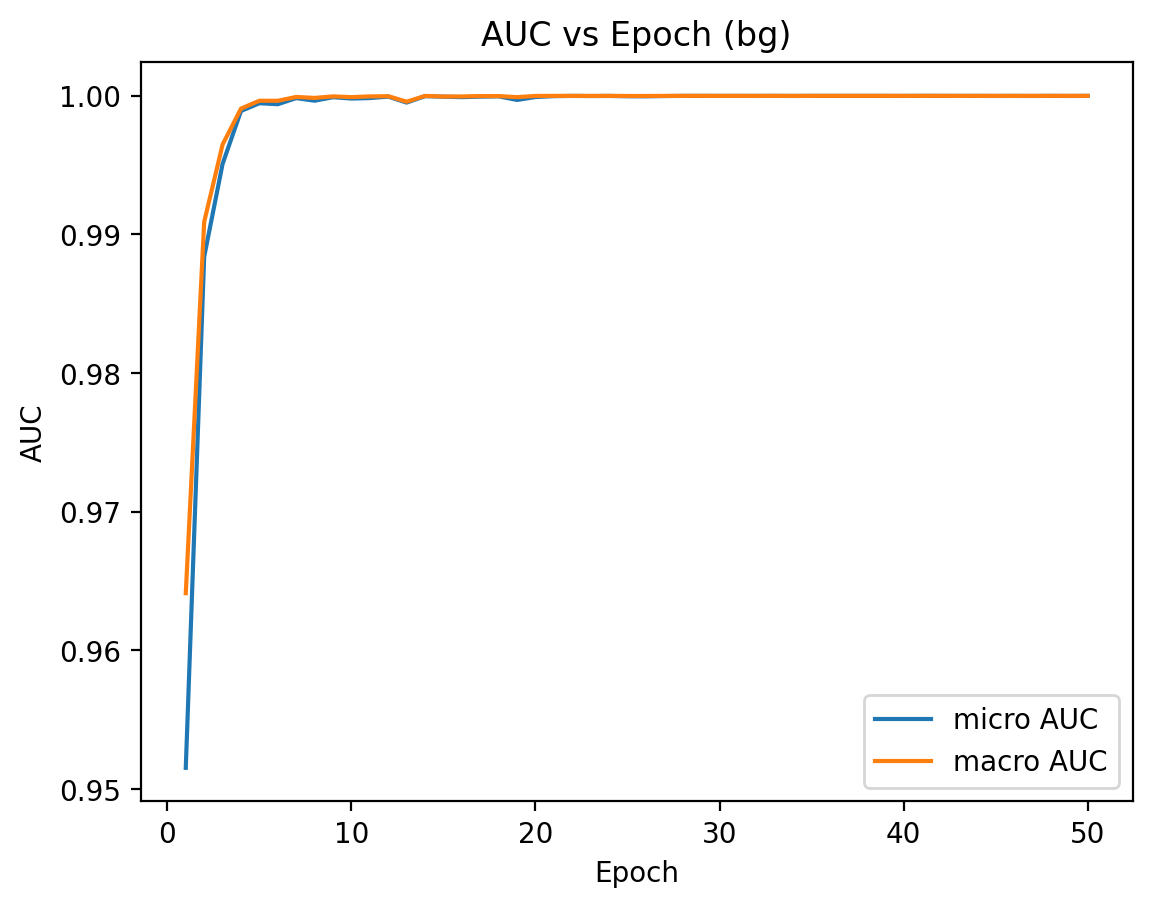}
    \caption*{\small BG}
  \end{subfigure}\hfill
  \begin{subfigure}[t]{0.32\textwidth}
    \centering
    \includegraphics[width=\linewidth]{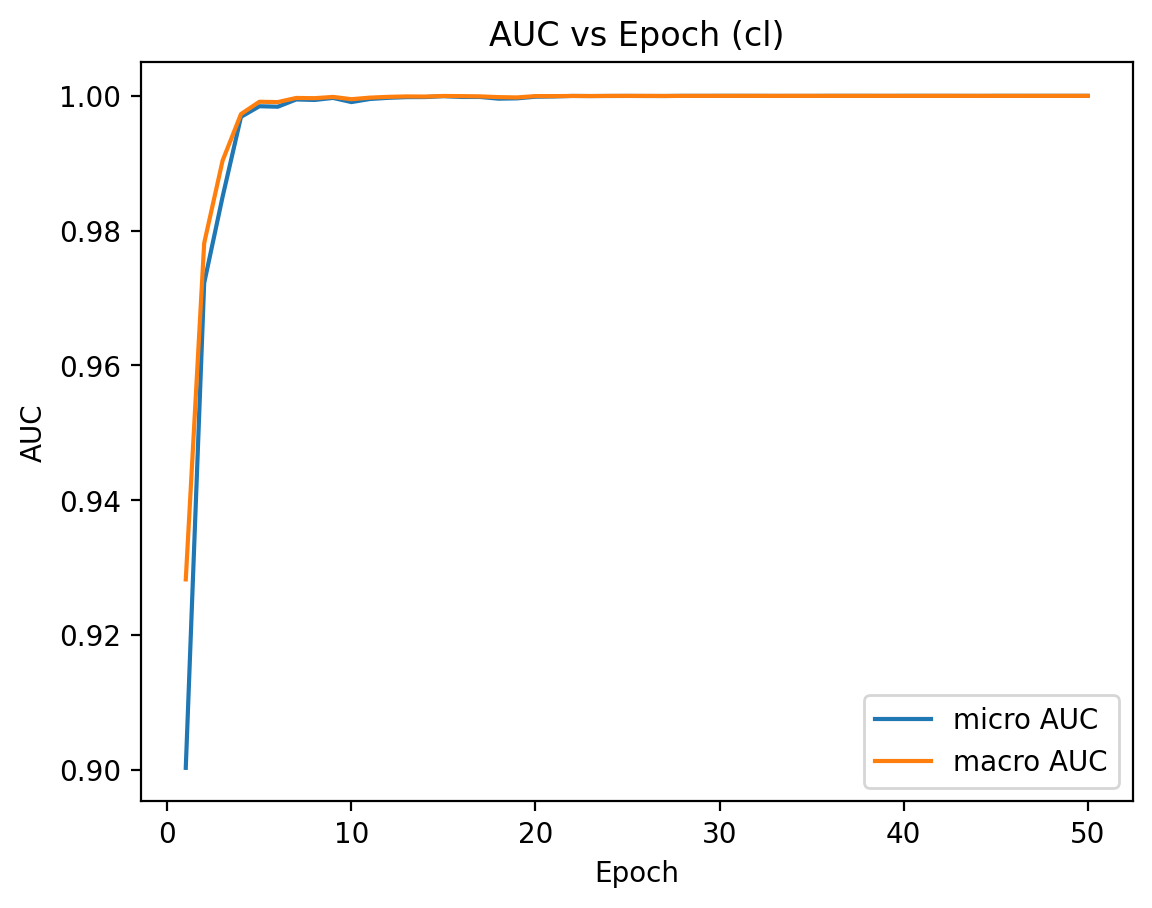}
    \caption*{\small CL}
  \end{subfigure}
  \caption{ROC and AUC for CNN+TKAN on CASIA--B, showing both micro- and macro-averaged one-vs-rest ROC curves, reported separately for NM, BG, and CL.}
  \label{fig:auc_curves}
\end{figure*}

The AUC-versus-epoch trends in \cref{fig:auc_curves} support the same conclusion: NM and BG remain near unity throughout training, while CL shows a steady and consistent advantage. Finally, \Cref{fig:curves} presents the training and validation trajectories of the CNN+TKAN model. Both loss and accuracy evolve smoothly and reach a stable plateau, with no visible indication of substantial late-stage overfitting. Together with the AUC trends, these observations suggest that TKAN improves both final recognition accuracy and ranking robustness without modifying the backbone architecture.

\FloatBarrier

% =========================================================
% 5. Conclusion
% =========================================================
\section{Conclusion}
This study presented a silhouette-based gait recognition framework that combines a compact CNN encoder with a Temporal Kolmogorov--Arnold Network (TKAN) for sequence modeling. The goal was to test whether a compact and efficient temporal design can improve performance under realistic surveillance conditions without increasing backbone complexity. On CASIA-B, the proposed CNN+TKAN model achieved the highest Rank-1 accuracy under the NM, BG, and CL evaluation settings, while the reported AUC trends indicated strong separability across decision thresholds. These findings suggest that improved temporal modeling, implemented through edge-wise learnable functions and a two-level memory mechanism, can enhance robustness to covariate variation while preserving a lightweight encoder.

The proposed framework is particularly relevant in settings where facial cues are unreliable, video clips are short or noisy, and computational resources are limited. At the same time, several limitations remain. The current formulation operates on fixed 50-frame clips, depends on silhouette quality, and is trained in a supervised closed-set setting, which may restrict generalization to unseen identities or deployment conditions.

Future work will examine streaming and variable-length inference, cross-dataset and in-the-wild evaluation, multimodal fusion with complementary cues such as pose or optical flow, and self-supervised or open-set learning strategies for improved generalization. Further analysis of the learned univariate functions within TKAN may also improve interpretability and support the development of even more compact variants.

% --- Flush floats before references but don't force a new page; balance columns
\FloatBarrier
% \balance

% =========================================================
% References
% =========================================================

\end{document}